\documentclass[10pt,twocolumn,letterpaper]{article}

\usepackage[pagenumbers]{cvpr} %

\usepackage[dvipsnames]{xcolor}
\usepackage{graphicx}
\usepackage{ifthen}

\usepackage{amsmath}
\usepackage{amssymb}
\usepackage{adjustbox}
\usepackage{caption}
\usepackage{calc}
\usepackage{booktabs}
\usepackage{colortbl}
\usepackage{tabularx}
\usepackage{tabu}
\usepackage{multirow}
\usepackage{array}
\usepackage{tikz,pgfplots}
\usepackage{amssymb}%
\usepackage{pifont}%
\usepackage{xspace}

\usetikzlibrary{arrows.meta, positioning, calc, shadows}

\newboolean{showcomments}
\setboolean{showcomments}{false} %

\newcommand{\name}{ControlRoom3D}

\newcommand{\cmark}{\ding{51}}%
\newcommand{\xmark}{\ding{55}}%
\definecolor{arrowgray}{RGB}{150,150,150}
\definecolor{profgreen}{RGB}{0,128,0}
\definecolor{darkgreen}{RGB}{0,255,0}
\definecolor{brightpurple}{RGB}{220,110,200}
\definecolor{kevincolor}{RGB}{147,112,219}

\newcommand\blfootnote[1]{%
  \begingroup
  \renewcommand\thefootnote{}\footnote{#1}%
  \addtocounter{footnote}{-1}%
  \endgroup
}

\newcommand{\parag}[1]{\vskip2pt \noindent \textbf{#1}}

\definecolor{m_red}{RGB}{255,209,209}
\definecolor{m_red_border}{RGB}{215,23,20}

\definecolor{m_orange}{RGB}{255,230,204}
\definecolor{m_orange_border}{RGB}{215,155,0}

\definecolor{m_blue}{RGB}{217,232,251}
\definecolor{m_blue_border}{RGB}{107,141,190}

\definecolor{m_yellow}{RGB}{255,242,205}
\definecolor{m_yellow_border}{RGB}{213,182,82}

\definecolor{m_violet}{RGB}{225,213,231}
\definecolor{m_violet_border}{RGB}{150,115,166}

\definecolor{m_green}{RGB}{213,232,212}
\definecolor{m_green_border}{RGB}{130,179,102}

\definecolor{m_gray}{RGB}{245,245,245}
\definecolor{m_gray_border}{RGB}{102,102,102}

\DeclareRobustCommand{\colorsquare}[1]{\tikz{\path[draw=#1_border,fill=#1, thick, rounded corners=0.6pt] (0,0) rectangle (6pt,6pt);}}

\DeclareRobustCommand{\colordot}[1]{%
    \tikz[baseline=-0.5ex]{
        \node[shape=circle, draw=#1_border, fill=#1, inner sep=2pt] (char) {};
    }%
}

\newlength{\myXheight}
\newcommand{\proxyroom}{$\mathcal{M}_\text{\includegraphics[trim={50pt 50pt 50pt 50pt},clip,height=\myXheight]{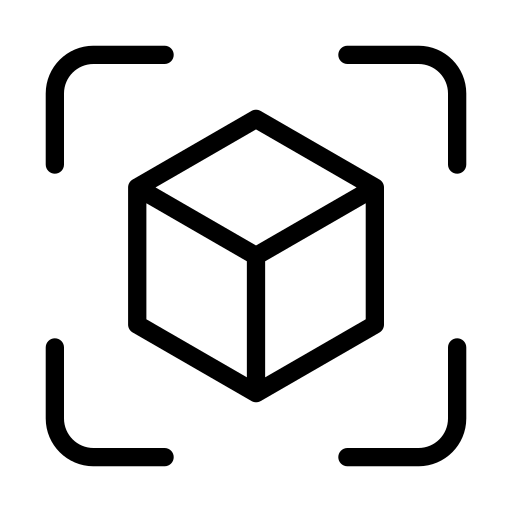}}$}

\newlength{\mybigXheight}
\def\eg{\emph{e.g.}\@\xspace} 
\def\ie{\emph{i.e.}\@\xspace} 
\def\cf{\emph{cf.}\@\xspace}

\definecolor{cvprblue}{rgb}{0.21,0.49,0.74}
\usepackage[pagebackref,breaklinks,colorlinks,citecolor=cvprblue]{hyperref}

\def\paperID{130} %
\def\confName{CVPR}
\def\confYear{2024}

\title{\vspace{-20px}\name{}: Room Generation using Semantic Proxy Rooms\vspace{-10px}}

\author{
\normalsize
Jonas Schult$^{1,2*}$ \hspace{8px}
Sam Tsai$^{1}$ \hspace{8px}
Lukas Höllein$^{1,3*}$ \hspace{8px}
Bichen Wu$^{1}$ \hspace{8px}
Jialiang Wang$^{1}$ \hspace{8px}
Chih-Yao Ma$^{1}$ \hspace{8px}
Kunpeng Li$^{1}$ \\
\normalsize
Xiaofang Wang$^{1}$ \hspace{8px}
Felix Wimbauer$^{1,3*}$ \hspace{8px}
Zijian He$^{1}$ \hspace{8px}
Peizhao Zhang$^{1}$ \hspace{8px}
Bastian Leibe$^{2}$ \hspace{8px}
Peter Vajda$^{1}$ \hspace{8px}
Ji Hou$^{1}$
\\
{\small
$^1$Meta GenAI
\hspace{15px}
$^2$RWTH Aachen University
\hspace{15px}
$^3$Technical University of Munich
}
\vspace{-25px}\\
\\\\
}

\begin{document}

\twocolumn[{%
\renewcommand\twocolumn[1][]{#1}%
\maketitle
\vspace{-25pt}
\includegraphics[width=\textwidth]{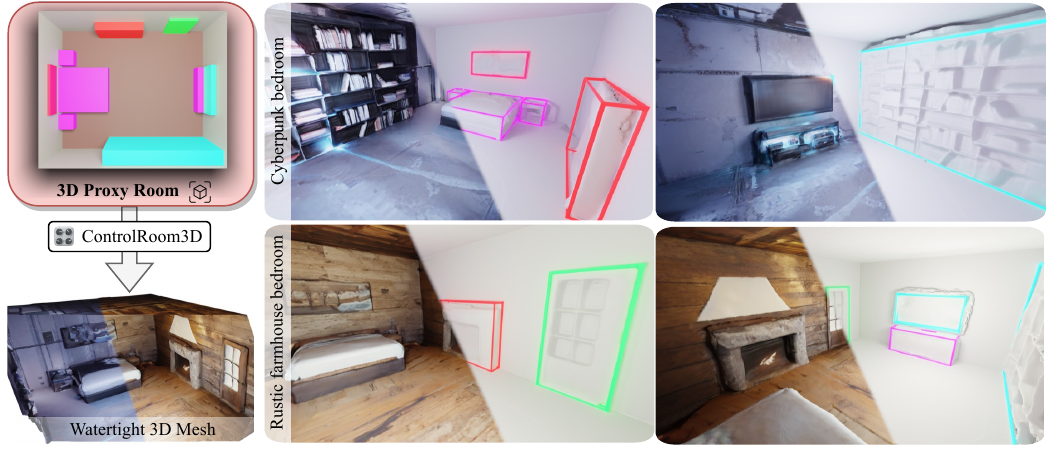}
\vspace{-23pt}
\captionof{figure}{\textbf{Textured 3D Mesh Generation from 3D Semantic Layouts and Text Prompts.}
Given a textual description of the overall room style and a rough 3D room layout based on 3D semantic bounding boxes, our method called \name{} creates diverse and globally plausible 3D room meshes which align well with the room layout.
Project Page: { \href{https://jonasschult.github.io/ControlRoom3D/}{jonasschult.github.io/ControlRoom3D/}}
}
\label{fig:teaser}
\vspace{10px}
}]
 
\maketitle

\begin{abstract}
\vspace{-10px}
\blfootnote{$^{*}$ Work performed during an internship at Meta GenAI.}
Manually creating 3D environments for AR/VR applications is a complex process requiring expert knowledge in 3D modeling software.
Pioneering works facilitate this process by generating room meshes conditioned on textual style descriptions. %
Yet, many of these automatically generated 3D meshes do not adhere to typical room layouts, compromising their plausibility, \eg, by placing several beds in one bedroom. %
To address these challenges, we present \emph{\name{}}, a novel method to generate high-quality room meshes.
Central to our approach is a user-defined 3D semantic proxy room that outlines a rough room layout based on semantic bounding boxes and a textual description of the overall room style.
Our key insight is that when rendered to 2D, this 3D representation provides valuable geometric and semantic information to control powerful 2D models to generate 3D consistent textures and geometry that aligns well with the proxy room.
Backed up by an extensive study including quantitative metrics and qualitative user evaluations, our method generates diverse and globally plausible 3D room meshes, thus empowering users to design 3D rooms effortlessly without specialized knowledge.

\end{abstract}    
\section{Introduction}
\vspace{-10px}
\label{sec:intro}
Generating high-quality mesh representations of 3D scenes is essential for creating immersive experiences in AR/VR applications.
However, manually synthesizing these 3D environments is a complex task that requires expertise in 3D modeling software~\cite{maya19,Hess2010blender}, posing a substantial barrier for end-users looking to create personalized scenes.
Fueled by the recent progress in 2D generative models~\cite{sohldickstein15icml,rombach22cvpr,ho20neurips,dhariwal21neurips}, an emerging line of work aims to simplify the creation process of 3D objects~\cite{wang23neurips,lin23cvpr,chen2023it3d,wang23sjc,tang2023dreamgaussian} or even room-scale scenes~\cite{hoellein23iccv,zhang23arXiv,fridman23arxiv,cohenbar2023arxiv}.
Among them, the recent work of Text2Room~\cite{hoellein23iccv} creates compelling textured 3D room-scale meshes using a text prompt describing the scene and an off-the-shelf text-to-image model~\cite{rombach22cvpr}. This is achieved by iteratively inpainting missing 2D regions of the rendered 3D mesh from a novel viewpoint.
Subsequently, depth is estimated for the new content, followed by backprojection and fusion with the current 3D mesh.
Despite Text2Room's ability to generate locally convincing meshes with detailed textures, it suffers from two crucial shortcomings due to its iterative mesh inpainting strategy solely based on local context:
First, Text2Room does not follow conventional room layouts, %
yielding unrealistic spatial arrangements of objects and walls.
Secondly, it tends to duplicate the most prominent objects of a scene type, \eg, placing several beds into a single bedroom, leading to semantically implausible scenes.

To address these shortcomings, we propose \emph{\name{}}, which generates diverse and globally plausible 3D room meshes.
A central element of our method is the 3D semantic proxy room (Fig.~\ref{fig:teaser},~\colorsquare{m_red}) which allows the user to specify semantic bounding boxes and textual descriptions of the desired room style that then serve as a rough outline for the generated room layout.
Our key insight is that this 3D representation, when rendered to 2D, provides valuable geometric and semantic information. This information can be used to control powerful 2D generative models, enabling the creation of 3D consistent textures and geometry that align with the proxy room.
However, projecting 3D controls to 2D comes with a loss of information which needs to be properly addressed.
As naively generating 2D views \emph{individually} is detrimental for scene style consistency, we first present a novel panorama generation method, which generates a comprehensive 360$^\circ$ view of the room.
Leveraging homographic constraints between equirectangular patches, we equip a latent diffusion model with correspondence-aware attention modules~\cite{tang23neurips} which establish a coherent and seamless style, while faithfully adhering to the projected semantics of the user-provided 3D proxy room.
Next, the quality of depth estimation becomes pivotal for converting this panorama into a 3D representation that aligns with the proxy room.
Building upon our insight that the 3D proxy room offers not only semantic but also valuable geometric information, we introduce a novel geometry alignment approach which leverages the spatial dimensions of 3D bounding boxes in an iterative optimization strategy to support a metric depth estimation network~\cite{bhat23arxiv} to generate 3D structures which align with the proxy room.
Thirdly, while panoramic views are valuable, they are insufficient for a complete 3D scene.
To address issues like occlusion and low-resolution textures, our method first eliminates low density regions, then uses new viewpoints for inpainting, guided by the proxy room and our geometry alignment, thus integrating new geometry into the mesh seamlessly.

To summarize, our contributions are:
\textbf{(1)} We propose \emph{\name{}} which generates diverse and semantically plausible 3D room meshes aligning well with the user-defined room layout and textual description of the room style.
\textbf{(2)} Leveraging our key insight that this room layout provides both valuable semantic and geometric priors, we present technical components including the semantic proxy room, guided panorama generation, mesh completion, and geometry alignment to enable coherent and seamless 3D textures and geometry.
\textbf{(3)} Using both quantitative and qualitative metrics, we validate the effectiveness of each component introduced and show that our method generates more plausible 3D rooms compared to competing methods.

\begin{figure*}[th!]
\centering
\includegraphics[width=1.0\textwidth]{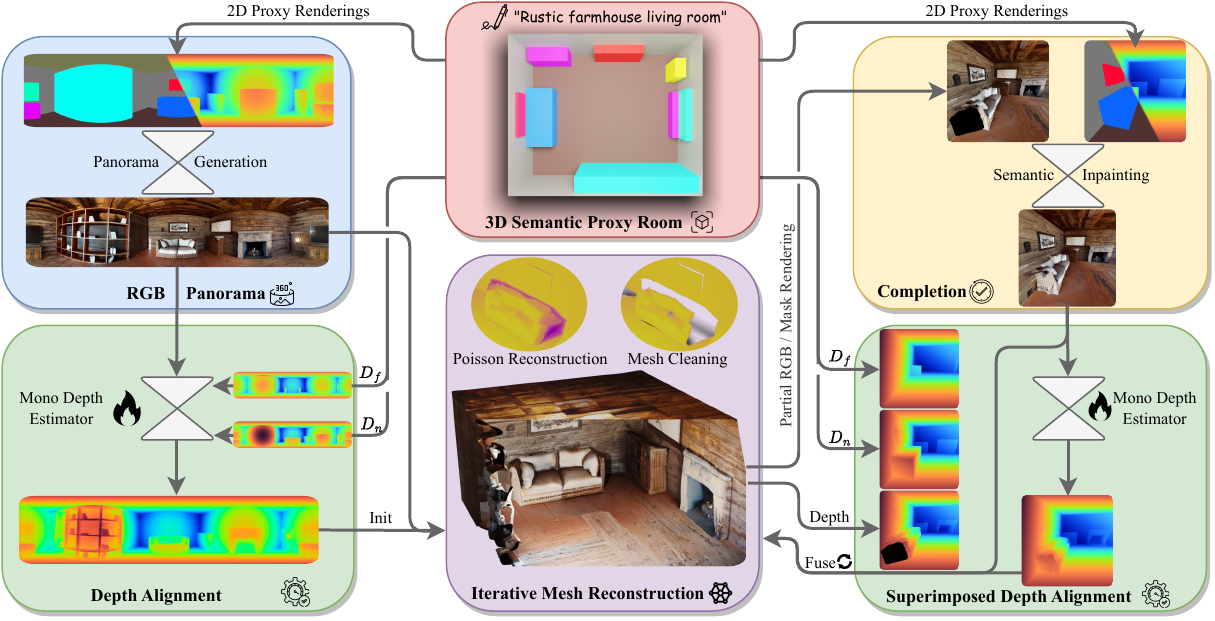}
\vspace{-20px}
\caption{\textbf{Method Overview.}
The \emph{3D semantic proxy room}~\colorsquare{m_red} contains a user-defined 3D room layout based on semantic bounding boxes and a textual description of the overall room style.
From this representation, we derive control signals that ensure 3D consistent textures and align corresponding depth estimates with the room layout.
The \emph{panorama generation module}~\colorsquare{m_blue} generates a comprehensive 360$^{\circ}$ view of the room, the depth of which is then estimated and aligned with the proxy room~\colorsquare{m_green}.
After Poisson reconstruction and artifact cleaning~\colorsquare{m_violet}, we update the room mesh by iteratively completing partial renderings of the scene~\colorsquare{m_yellow} and estimating the depth of newly added textures~\colorsquare{m_green}.
\vspace{-10px}
}
\label{fig:method_overview}
\end{figure*}
\section{Related Work}
\parag{Text-to-3D.}
Fueled by the availability of large-scale vision-text datasets~\cite{schuhmann2021laion,yu2023scaling,schuhmann2022laion,pham2023combined,jia2021scaling,thomee2016yfcc100m,sharma18acl} and powerful vision-language models~\cite{radford2021learning}, recent advancements in 2D diffusion models~\cite{dalle3,rombach22cvpr,dai23arxiv,ramesh22arxiv,saharia22neurips,nichol21arxiv} have significantly contributed to the ongoing progress in generating 3D objects and scenes.
In particular, one prominent research direction focuses on single object synthesis~\cite{tang2023dreamgaussian,muller23cvpr,wang23neurips,poole23iclr,lin23cvpr,zhang23arXiv}, wherein a differentiable 3D representation~\cite{mildenhall2020nerf,kerbl3Dgaussians} is commonly optimized through a loss signal derived from the denoising of rendered views~\cite{wang23sjc,poole23iclr,wang23neurips}.
Despite achieving impressive results in the synthesis of object-centric scenes, the challenge of generalizing these techniques to large and complex scenes persists~\cite{wang23neurips}.
Typically, existing approaches rely on additional 3D data sources~\cite{devries21iccv, bautista22neurips, bahmani23iccv}, which are challenging to obtain at scale.
Two notable approaches related to our work include SceneScape~\cite{fridman23arxiv} and Text2Room~\cite{hoellein23iccv}.
These methods leverage off-the-shelf text-conditioned inpainting models~\cite{rombach22cvpr} in combination with state-of-the-art monodepth estimation models~\cite{bae22bmvc, ranftl20tpami} to create a 3D mesh representation.
Alternatively, LDM3D~\cite{stan23arxiv} directly predicts an RGBD frame based on a given text prompt.
However, these approaches have limited control over the generation process by only being able to change the text prompt at each inpainting step, typically leading to globally implausible room layouts.
In contrast, \name{} allows precise control over the scene composition by using globally consistent 2D maps derived from 3D semantic proxy rooms to guide the generation process. %

\parag{Semantically Guided 3D Scene Generation.}
An emerging line of works investigates scene generation with semantic compositing~\cite{gafni22eccv, esser20cvpr,rombach22cvpr,chen23arxiv,xie23iccv,li23cvpr,yang23cvpr}. 
However, in 3D they are typically constrained to low resolution textures and partial room settings to enable inward-facing views~\cite{cohenbar2023arxiv, po2023arxiv,zhang23arXiv} or require synthetic 3D datasets and thus are restricted to the room style present in the dataset~\cite{bahmani23iccv}.
At the same time, current diffusion models show great capabilities to be extended by control signals such as pixel-perfect maps, keypoints or bounding boxes~\cite{zhang23iccv,chen23arxiv,xie23iccv,li23cvpr,yang23cvpr}.
We follow this line of work and show that 2D control signals derived from a 3D representation can be used to produce 3D consistent high-resolution images of large diversity without the need of inward-facing views.

\parag{Mesh Texturization.}
A related line of work deals with re-texturizing a given mesh representation of 3D scenes~\cite{song23arxiv,tang23neurips} or objects~\cite{michel22cvpr} using a stylization text prompt.
Unlike \name{}, these methods focus on re-texturizing the mesh in a novel style without (or only minor adaptations) to the original geometry while our approach not only generates stylized textures but also creates novel geometry \emph{from scratch} which aligns with the style of the texture.

\section{Method}
\vspace{-5px}
\parag{Overview.}
(Fig.~\ref{fig:method_overview}) Given a textual description of the overall room style and a rough 3D room layout based on 3D semantic bounding boxes, \name{} creates diverse and globally plausible 3D room meshes which align well with the room layouts.
The key technical components of our method comprise the \emph{proxy room}~\colorsquare{m_red}, from which we derive 2D semantic and guidance depth maps for arbitrary viewpoints in the scene; the \emph{geometry alignment module}~\colorsquare{m_green}, to align newly generated 2D views to the proxy room; the \emph{panorama generation module}~\colorsquare{m_blue}, to initially generate a globally style-consistent 360$^\circ$ view of the scene; and a final \emph{completion module}~\colorsquare{m_yellow}, which samples novel camera viewpoints to inpaint occluded and unobserved regions in the mesh with high-resolution textures.
Details of each of these steps are discussed in the following sections, and we highlight each component with its corresponding color of Fig.~\ref{fig:method_overview}.

\parag{3D Semantic Proxy Room.}
\begin{figure}[t]
\includegraphics[width=\linewidth,trim={0cm 0cm 0cm 0cm},clip]{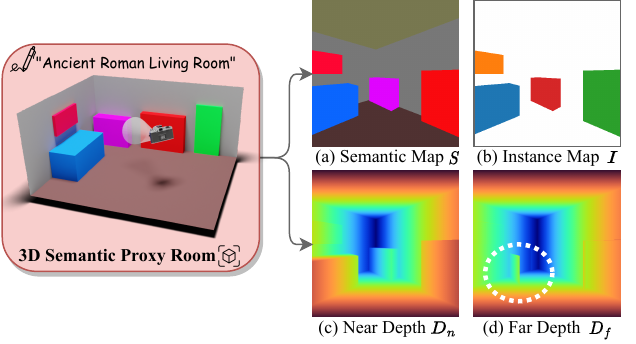}
\vspace{-20px}
\caption{\textbf{Proxy Room Renderings.}
We use classical rasterization to render semantic segmentation and instance maps as well as near and far depth maps from the proxy room.
\vspace{-10px}}
\label{fig:proxy_room_renderings}
\end{figure}
(Fig.~\ref{fig:method_overview},~\colorsquare{m_red}) At the core of our method lies the user-defined 3D semantic proxy room, which acts as a rough outline for the room layout and is defined by semantic bounding boxes and a textual description of the desired room style.
Specifically, we define a proxy room \proxyroom\ as a set of semantic bounding boxes $\mathcal{B}=(p,s,c,i)$, where $p \in \mathbb{R}^3$ is the box center, $s \in \mathbb{R}^3$ the box size, $c$ the semantic class id, and $i$ the unique instance id associated with each box.
Our key insight is that this 3D representation, when rendered to 2D, provides valuable geometric and semantic information that can be used to control powerful 2D generative models, enabling the creation of 3D consistent textures and geometry that align with the proxy room.
We render the proxy scene from a given viewpoint $V$, producing multiple types of frames:
$$S, I, D_n, D_f = r(\mathcal{M}_\text{\includegraphics[trim={50pt 50pt 50pt 50pt},clip,height=\myXheight]{fig/img/proxy_symbol.png}}, V)$$
Here, $r(\cdot)$ is standard rasterization without shading, the outputs $S$$\in$$[1,C]^{H \times W}$, $I$$\in$$\mathcal{N}^{H \times W}$, $D_n, D_f$$\in$$\mathbb{R}_{+}^{H \times W}$ are the rendered 2D semantic segmentation map of $C$ classes, the instance segmentation map, and the near and far guidance depth map, respectively (Fig.~\ref{fig:proxy_room_renderings}).
In the following, we detail how these rendered maps are utilized in our components to achieve room mesh generation that aligns both semantically and geometrically with the proxy room.

\parag{Guided Panorama Generation.}
\begin{figure}[t]
\centering
\includegraphics[width=\linewidth]{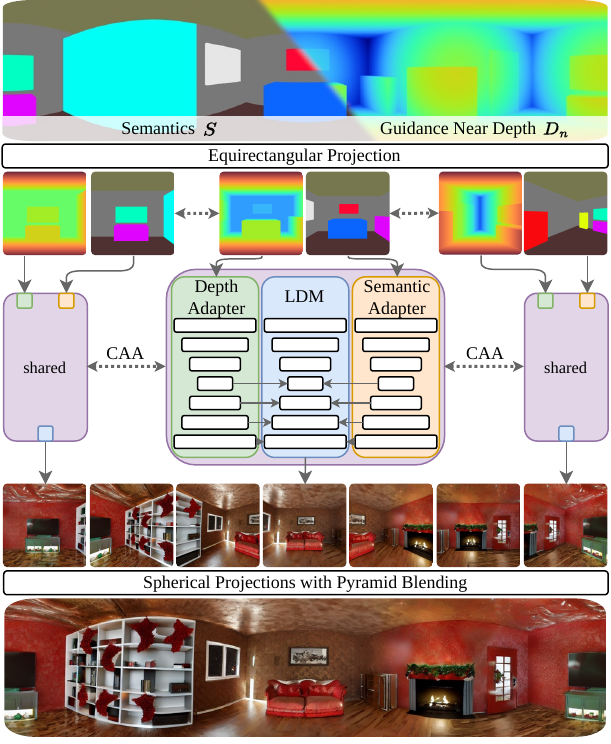}
\vspace{-20px}
\caption{\textbf{Guided Panorama Generation.}
We generate 360$^\circ$ panoramas guided by 3D semantic bounding boxes rendered from the proxy room.
To this end, we equip an LDM~\colorsquare{m_blue} with correspondence-aware attention modules (CAA)~\cite{tang23neurips} and semantic~\colorsquare{m_orange} and depth~\colorsquare{m_green} adapters finetuned on semantic bounding boxes.
\vspace{-25px}
}
\label{fig:sem_mvdiffusion}
\end{figure}
(Fig.~\ref{fig:method_overview},~\colorsquare{m_blue} and Fig.~\ref{fig:sem_mvdiffusion})
We observe that naively generating 2D views \emph{individually} or only relying on iterative inpainting is detrimental for scene style consistency as the overall style tends to drift with each generated image and inaccurate loop closure~\cite{tang23neurips}.
Instead, we jointly create a comprehensive 360$^\circ$ view of the room that establishes a coherent and seamless style.
This process is guided by the 3D semantic bounding boxes projected from the proxy room (Fig.~\ref{fig:proxy_room_renderings}\textcolor{red}{a,c}).
To harness readily available powerful 2D generation models trained on perspective images, we first divide the semantic panorama into 8 views, each encompassing a 90$^\circ$~FOV and a 45$^\circ$ horizontal rotation using equirectangular projections, providing pixel-perfect correspondences between adjacent frames defined by $3$$\times$$3$ homography matrices.
We then equip a latent diffusion model (LDM) with 3 types of modules.
Firstly, correspondence-aware attention modules~\cite{tang23neurips} use homographic correspondences between adjacent frames.
Secondly, a semantic adapter guides the generation towards the specified room layout, while thirdly, the depth adapter conditions the generation on the guidance depth map.
We employ pyramid blending to create the final panoramic image from the spherical projections of the generated images~\cite{tang23neurips}.

\parag{Adapter Fine-tuning on HyperSim Bounding Boxes.}
Off-the-shelf semantic adapters gain strong semantic understanding through training on large datasets, but they are typically trained on pixel-perfect semantic masks.
This stands in stark contrast to our control input, which involves 3D bounding boxes rendered in 2D views that only provide a rough outline of the object (Fig.~\ref{fig:proxy_room_renderings}\textcolor{red}{a}).
To bridge this gap, we construct a dataset derived from HyperSim~\cite{roberts21iccv}, containing photo-realistic renderings of indoor environments with 3D semantic bounding box annotations.
We render 3D bounding boxes for each scene and fine-tune the semantic and depth adapters on this dataset.
Additional details can be found in our supplementary material.
\begin{figure}[t]
\includegraphics[width=\linewidth,trim={0cm 0cm 0cm 0cm},clip]{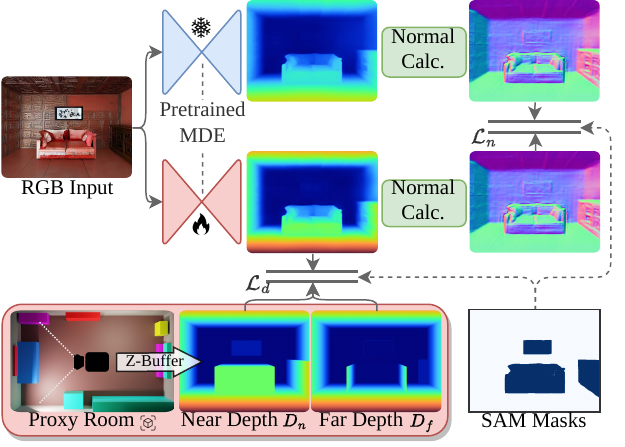}
\vspace{-20px}
\caption{\textbf{Geometry Alignment.}
We align generated 2D textures with the proxy room~\colorsquare{m_red}.
To achieve this, we derive both near and far depth maps by rendering from an identically posed camera inside the proxy room and align the generated objects (extracted with SAM~\cite{kirillov23arxiv}) accordingly by fine-tuning a pre-trained depth estimation network.
Additionally, we preserve the overall object shape by ensuring the integrity of the normals during optimization.
\vspace{-15px}
}
\label{fig:depth_alignment}
\end{figure}
\parag{Geometry Alignment.}
(Fig.~\ref{fig:method_overview},~\colorsquare{m_green} and Fig.~\ref{fig:depth_alignment})
After generating panoramas, the next step is to transform these 2D textures into a 3D representation by predicting per-pixel depth.
Naively estimating depth using a standard depth estimator can yield subpar results due to scale ambiguity. 
However, the proxy room offers valuable geometric information, including the near and far guidance depth map (Fig.~\ref{fig:proxy_room_renderings}\textcolor{red}{c,d}).
A fundamental aspect of \name{} is to align the predicted depth with this geometric prior.
Specifically, we optimize the depth predicted by a pre-trained metric depth estimator (MDE)~\cite{bhat23arxiv} to fit within its respective bounding boxes while preserving the object's shape (Fig.~\ref{fig:depth_alignment}).
To achieve this, we introduce a geometry alignment loss:
\begin{equation*}
\label{eq:depth_align_eq}
    \mathcal{L}_{d} = 
    \begin{cases}
        0 & \text{if } D_n \leq \hat{d} \leq D_f, \\
        \min\left( \lVert\hat{d} - D_n \rVert_1 , \lVert\hat{d} - D_f\rVert_1 \right) & \text{otherwise.}
    \end{cases}
\end{equation*}
Here, $\hat{d}$ is the predicted depth, and $D_n, D_f$ denote the near and far depth values.
Note that for walls, ceiling, and floor $D_n$$=$$D_f$.
This loss penalizes depth predictions that fall outside the designated bounding box with an $L_1$ loss.
Additionally, we employ SAM~\cite{kirillov23arxiv} to extract pixel-perfect instance masks for each generated instance by prompting it with the bounding box extracted from the instance maps $I$ (Fig.~\ref{fig:proxy_room_renderings}\textcolor{red}{b}).
For pixels within the rendered bounding box but outside the SAM mask, we set $D_n$ to $D_f$.
During test time, we optimize a pre-trained MDE using this loss, thereby aligning the depth with the proxy room while retaining the depth estimator's strong prior learned from large datasets.
While the depth alignment loss $\mathcal{L}_d$ aligns the frame with the proxy room, it may alter the object's surface significantly while fitting it into its bounding box.
To mitigate this, we present a normal preservation loss:
$\mathcal{L}_{n} = \lVert\hat{n} - n_0 \rVert_1$
where $\hat{n}$ are the estimated normals from the current depth prediction $\hat{d}$ and $n_0$ are the normals from the original MDE.

Once we have predicted depth values that align with the proxy room, our next step is to convert the 2D textures into a 3D representation.
However, a straightforward unprojection using the depth values can result in visible seams in overlapping regions, because the depth values for each frame are individually optimized.
To address this challenge, we first convert the $z$-depth values into distance values, as they remain invariant to rotations.
We then combine all frames into a panoramic view using spherical projections scaled by their corresponding distance values followed by pyramid blending, ultimately yielding the initial 3D representation.

\begin{figure}[t]
\centering
    \centering
    \includegraphics[width=1.0\linewidth]{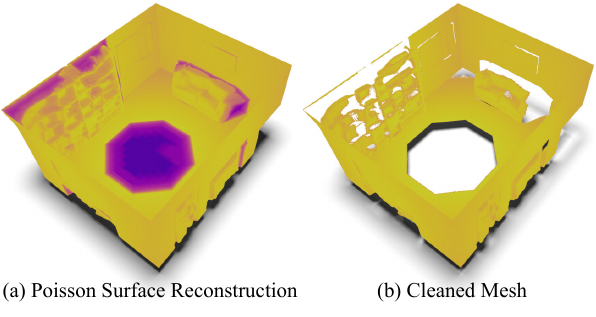}
    \vspace{-20px}
    \caption{\textbf{Mesh Cleaning.}
    After obtaining a watertight mesh with Poisson surface reconstruction (\emph{left}), we delete vertices whose densities fall below a threshold (density is depicted with darker color shades).
    After removing small connected components, we obtain a cleaned mesh ready to be inpainted with novel content from favourable viewpoints (\emph{right}).
    \vspace{-15px}
    }
    \label{fig:mesh_cleaning}
\end{figure}
\begin{figure}[t]
\centering
\includegraphics[width=\linewidth]{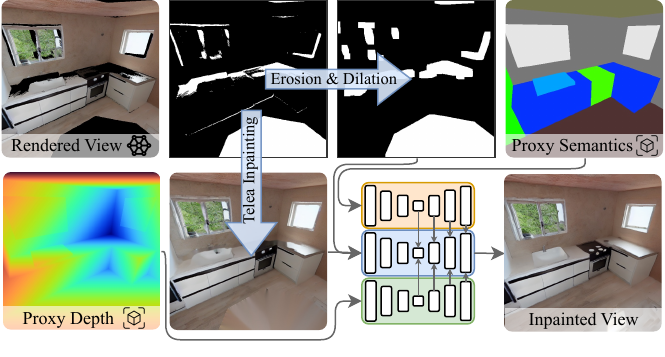}
\vspace{-20pt}
\caption{\textbf{Completion Pipeline.}
After inpainting small holes using classical Telea inpainting, we employ a 2D LDM~\colorsquare{m_blue} guided by a semantic~\colorsquare{m_orange} and depth~\colorsquare{m_green} adapter for larger missing areas, effectively inpainting these areas while aligning them to the control signals from the proxy room~\proxyroom{}.
}
\label{fig:completion_pipeline}
\end{figure}
\parag{Mesh Cleaning.}
(Fig.~\ref{fig:method_overview},~\colorsquare{m_violet} and Fig.~\ref{fig:mesh_cleaning})
While RGBD panoramas offer an initial 3D representation, they are susceptible to issues like noisy depth estimates and partially low-resolution textures, particularly in areas that are occluded or where mesh faces are stretched.
To address these challenges, our approach begins by identifying and filtering out regions of low quality in the mesh.
We then synthesize new content for these regions from a more favorable viewing angle.
To this end, we first perform Poisson surface reconstruction, converting the 3D point cloud obtained from the RGBD panorama into a watertight mesh representation $\mathcal{M}_\text{\includegraphics[trim={5pt 5pt 5pt 5pt},clip,height=\myXheight]{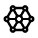}}$ (Fig.~\ref{fig:mesh_cleaning}\textcolor{red}{a}).
We proceed to remove mesh vertices and their adjacent faces if their density falls below a specified threshold $\delta_v$.
Finally, we eliminate floating artifacts that are disconnected from the main mesh by deleting components with a vertex count below a specified threshold $\delta_c$ (Fig.~\ref{fig:mesh_cleaning}\textcolor{red}{b}).
\parag{Mesh Completion.}
(Fig.~\ref{fig:method_overview},~\colorsquare{m_yellow} and Fig.~\ref{fig:completion_pipeline})
After Poisson reconstruction and artifact removal, we further refine the room mesh by iteratively inpainting the scene from various favorable viewpoints $V$, effectively filling missing areas in the mesh with novel content.
To this end, we render the current 3D mesh $\mathcal{M}_\text{\includegraphics[trim={5pt 5pt 5pt 5pt},clip,height=\myXheight]{fig/img/mesh_icon.jpg}}$ with classical rasterization $r(\cdot)$ to obtain the RGB frame $F$, depth map $D_\text{\includegraphics[trim={5pt 5pt 5pt 5pt},clip,height=\myXheight]{fig/img/mesh_icon.jpg}}$ as well as the image mask $M$ that identifies pixels without observed content:
$F, D_\text{\includegraphics[trim={5pt 5pt 5pt 5pt},clip,height=\myXheight]{fig/img/mesh_icon.jpg}}, M = r(\mathcal{M}_\text{\includegraphics[trim={5pt 5pt 5pt 5pt},clip,height=\myXheight]{fig/img/mesh_icon.jpg}}, V)$. 
In order to unleash the full potential of powerful 2D generation models, which perform better with larger connected regions, we first inpaint small gaps using traditional Telea inpainting~\cite{hoellein23iccv,fridman23arxiv,telea04jgt}.
We then expand the remaining holes through a sequence of morphological operations, specifically erosion followed by dilation.
This modified mask, combined with the rendered 2D semantic $S$ and guidance depth map $D_n$ from the proxy room, guides the inpainting of the missing areas using a semantic and depth adapter, yielding the final generated RGB frame.
We integrate this new content into the mesh using our geometry alignment module (Fig.~\ref{fig:method_overview},~\colorsquare{m_green}).
For seamless integration with pre-existing structures, we superimpose the rendered depth $D_\text{\includegraphics[trim={5pt 5pt 5pt 5pt},clip,height=\myXheight]{fig/img/mesh_icon.jpg}}$ on the near and far guidance depth map $D_n$, $D_f$ to optimize the depth of novel content to correspond to already generated 3D structures, \ie, $D^{i,j}_n=D^{i,j}_f=D^{i,j}_\text{\includegraphics[trim={5pt 5pt 5pt 5pt},clip,height=\myXheight]{fig/img/mesh_icon.jpg}}$ for $M^{i,j}=0$.
We repeat this process for a number of selected viewpoints, effectively inpainting any missing region while seamlessly integrating novel content.%

\section{Experiments}

\parag{Evaluation Setup and Metrics.}
We use distinct room layouts encompassing typical indoor rooms such as \emph{bedroom} or \emph{kitchen} but also unique settings such as \emph{wellness spa} with an indoor pool as targeted room for evaluation.
For text prompting, we define 12 room styles, spanning a diverse range of geographical (\eg, \emph{Japanese}, \emph{Scandinavian}), seasonal (\eg, \emph{Halloween}, \emph{Christmas}), historical (\eg, \emph{Ancient Roman}, \emph{Stone-age cave}), and artistic (\eg, \emph{cyberpunk, industrial}) themes.
We ask 12 participants to rate these scenes on a scale of 1--5 with respect to the following dimensions: 3D structure completeness (3DS), proxy alignment to the 3D guidance (PA), generated room layout plausibility (LP), and overall perceptual quality based on user preference (PQ).
We render 25sec long videos with 650 frames each for each generated room showing all objects from multiple (challenging) angles.
We let users rate each scene individually
(no head-to-head comparison).
In total, we collected 1120 data points using 175 scenes from the user study. %
We provide further details about the user study in the supplementary material.
Moreover, we calculate the CLIP score (CS)~\cite{hessel2021clipscore} and Inception score (IS)~\cite{salimans16nips} on 100 RGB renderings from random viewpoints for each scene.
\vspace{-10pt}

\paragraph{Implementation Details.}
We use a pre-trained LDM~\cite{rombach22cvpr} as well as semantic and depth adapter modules.
We finetune both the adapter weights on our 3D bounding box dataset derived from the HyperSim~\cite{roberts21iccv} dataset using BLIPv2 captions~\cite{li23icml}.
The correspondence-aware attention module (CAA) is trained on the Matterport3D~\cite{chang17matterport} dataset.
Our metric depth estimator is a pre-trained ZoeDepth~\cite{bhat23arxiv} model.
We use PyTorch3D~\cite{ravi2020pytorch3d} for mesh rasterization and fusion for the 3D proxy room as well as the final generated mesh.

\subsection{Comparison with State-of-the-Art Methods}

\parag{Baselines.}
There are no directly comparable methods which generate 3D textured meshes of entire rooms given a 3D room layout and a textual style description.
We thus adapt top-performing related methods for this task.
Text2Room~\cite{hoellein23iccv} uses a text-conditioned inpainting model followed by mono-depth inpainting~\cite{bae22bmvc}.
Following their original instructions, we enforce a room layout by changing the text prompt during the camera trajectory to reflect which objects are visible in the current frame.
MVDiffusion~\cite{tang23neurips} creates panorama images by enforcing correspondences between overlapping image crops using homography constraints.
Here, for each image crop we adjust the text condition to reflect which objects should be visible and use the same off-the-shelf mono-depth estimator~\cite{bhat23arxiv} and poisson mesh reconstruction as our method. %

\begin{figure*}[th!]
\centering
    \centering
    \includegraphics[width=1.0\textwidth]{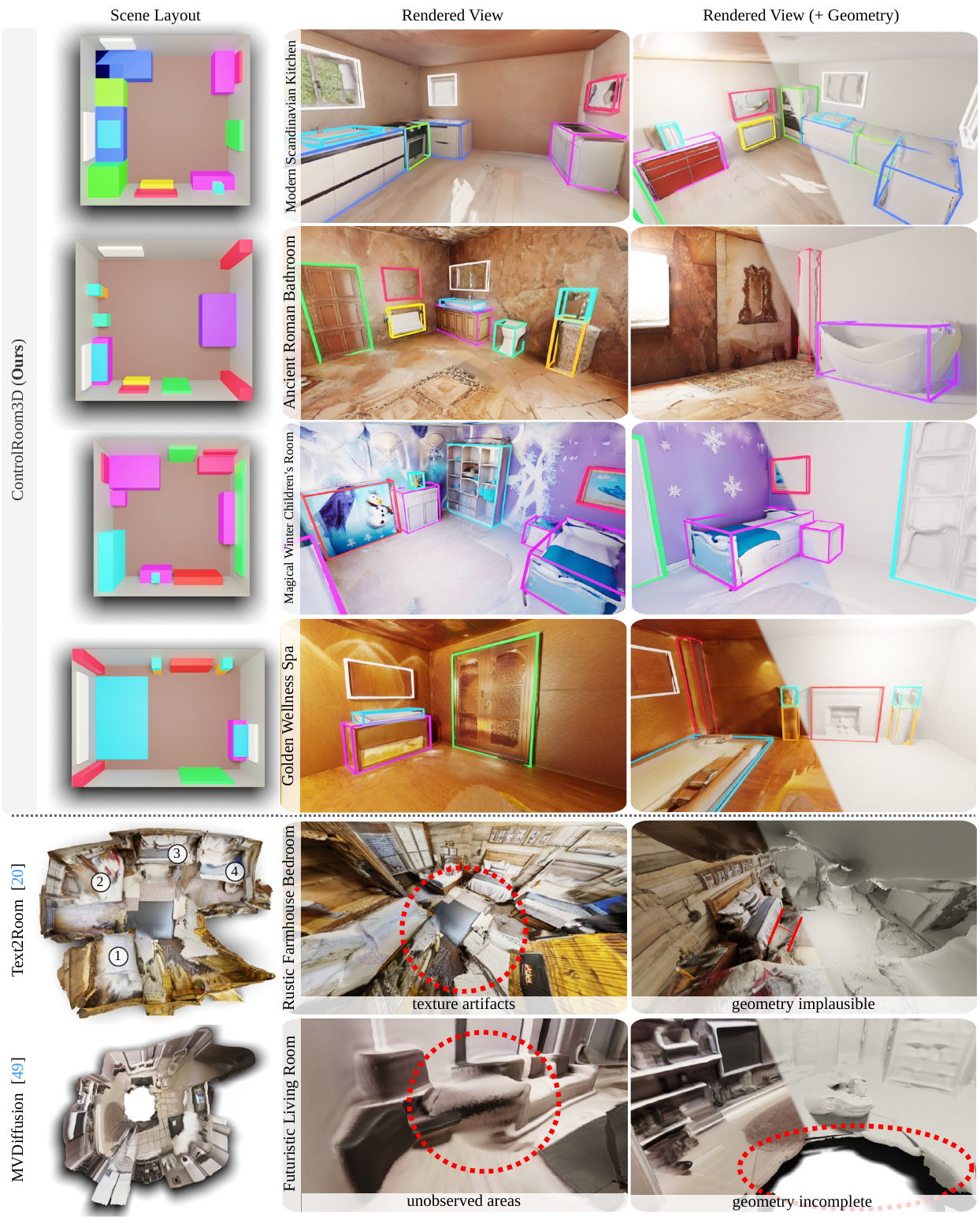}
    \vspace{-25px}
    \caption{\textbf{Qualitative Comparison with Baseline Methods.}
    We show colored geometry renderings of our method and two baseline methods.
    \name{} generates convincing geometries and textures given a user-defined room layout and a textual style description.
    We provide further example images and videos in the supplementary material.
    \vspace{-29px}
    }
    \label{fig:qualitative_comparison}
\end{figure*}

\parag{Results.}
\begin{table}
  \caption{
  \textbf{Quantitative Comparison.}
  We report 2D metrics and user study results, including: CLIP Score~(CS), Inception Score~(IS), 3D Structure Completeness~(3DS), Layout Plausibility~(LP) and Perceptual Quality~(PQ).
  }
  \centering
  \resizebox{\linewidth}{!}{
  \begin{tabular}{l cc ccc}
    \toprule
        \multirow{2}{*}{Method} & \multicolumn{2}{c}{2D Metrics} & \multicolumn{3}{c}{User Study \small{(3D Mesh)}}\\
                        \cmidrule(l{2pt}r{2pt}){2-3} \cmidrule(l{2pt}r{2pt}){4-6}
    & CS $\uparrow$ & IS $\uparrow$ & 3DS $\uparrow$ & LP $\uparrow$ & PQ $\uparrow$\\
    \midrule
    Text2Room~\cite{hoellein23iccv} & 24.0 & \textbf{6.47} & 1.71 & 2.12 & 2.33 \\
    MVDiffusion~\cite{tang23neurips} & 25.6 & 6.37 & 2.48 & 3.01 & 2.89 \\
\arrayrulecolor{black!10}\midrule\arrayrulecolor{black}
    \name{} (\textbf{Ours}) & \textbf{28.4} & 5.75 & \textbf{4.19} & \textbf{4.54} & \textbf{4.07}\\
    \bottomrule
  \end{tabular}
  }
  \label{tab:ours-baseline}
\end{table}
In our study, we provide a detailed quantitative comparison with top-performing methods in Tab.~\ref{tab:ours-baseline} and illustrate qualitative examples in Fig.~\ref{fig:qualitative_comparison}.
A critical finding is that text-only scene conditioning is inadequate for generating realistic room layouts.
Instead, our method integrates 2D semantic and geometric controls based on a 3D proxy room~\colorsquare{m_red} leading to more plausible layouts with an improvement of +1.5 LP. 
For instance, Text2Room frequently but incorrectly places multiple instances of the most prominent object of a scene type in the room, such as 4 beds in a single bedroom (Fig.~\ref{fig:qualitative_comparison}, \emph{bottom}), resulting in unrealistic layouts. 
In contrast, our approach accurately reflects the intended layout, marked by colored bounding boxes on the renderings.
This accuracy enables more efficient completion of missing areas 
reducing artifacts (+1.7 3DS) and enhancing overall perceptual quality (+1.2 PQ) while faithfully following the text conditioning (+2.8 CS). 
The inception score (IS) tends to favor methods that generate diverse scenes, even if they do not necessarily represent plausible room layouts (LP).

Nonetheless, our method displays versatility in creating a wide range of scene types and layouts, as shown in Fig.~\ref{fig:qualitative_comparison} and supplementary videos.

\begin{figure*}[t]
\centering
\includegraphics[width=\linewidth]{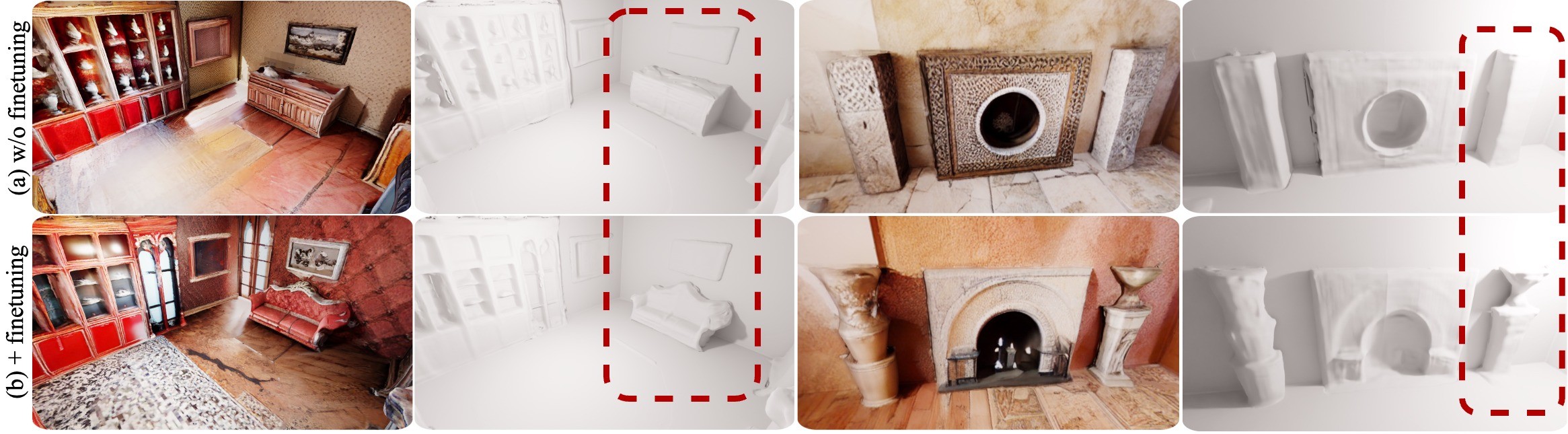}
\vspace{-20px}
\caption{\textbf{Fine-tuning with rendered bounding boxes from HyperSim.}
Prior to fine-tuning with HyperSim data, \name{} tends to create box-shaped objects and struggles to accurately generate objects within their respective bounding boxes (\emph{a}), whereas fine-tuning enables \name{} to fill 3D bounding boxes with appropriate content (\emph{b}).
\vspace{-10px}
}
\label{fig:hypersim_comparison}
\end{figure*}
\begin{figure*}[t]
\centering
\includegraphics[width=\linewidth]{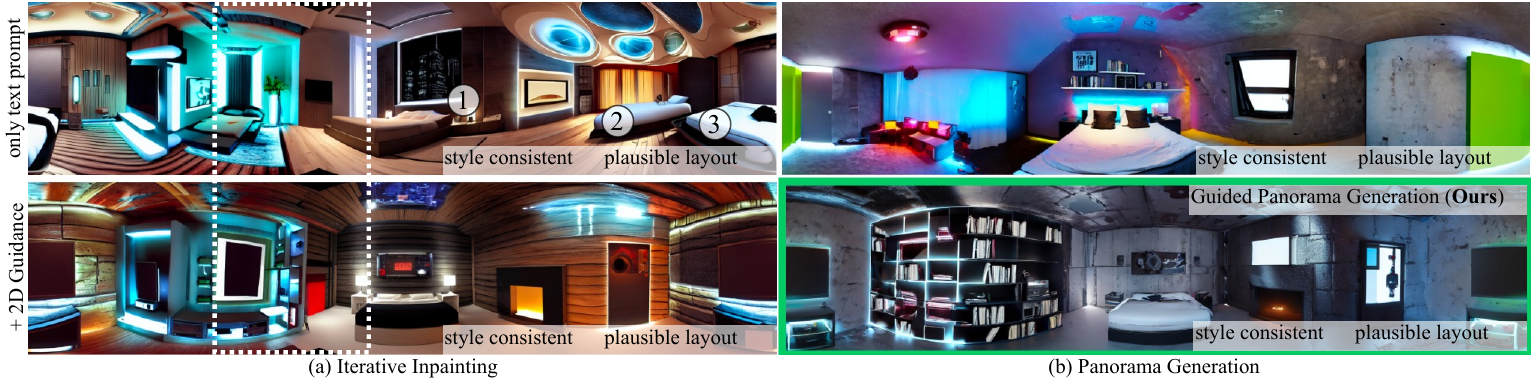}
\begin{tikzpicture}[overlay, remember picture]
    \tikzstyle{mynode} = [circle, fill=white, fill opacity=1.0, draw, thick, inner sep=1pt, minimum size=12pt]
    \node[above, text=red!60!black] at (-20.0mm, 7.4mm) {\xmark};
    \node[above, text=green!60!black] at (-1mm, 7.4mm) {\cmark};
    \node[above, text=red!60!black] at (-20.0mm, 27.7mm) {\xmark};
    \node[above, text=red!60!black] at (-1mm, 27.7mm) {\xmark};
    \node[above, text=green!60!black] at (66.0mm, 27.8mm) {\cmark};
    \node[above, text=red!60!black] at (84.5mm, 27.8mm) {\xmark};
    \node[above, text=green!60!black] at (65.5mm, 7.9mm) {\cmark};
    \node[above, text=green!60!black] at (84.5mm, 7.9mm) {\cmark};
\end{tikzpicture}
\vspace{-23pt}
\caption{\textbf{2D Guided Panorama Initialization.}
We find that only using text prompts to generate panoramas result in subpar layouts, particularly, iterative inpainting repeats the most prominent object in the scene (\emph{top, left}).
Adding 2D guidance with semantic and depth maps, results in more plausible room layouts, whereas panorama generation yields a better style consistency (\emph{bottom, right}).
\vspace{-15px}
}
\label{fig:sem_control_mvdiffusion}
\end{figure*}
\subsection{Analysis Experiments}
\begin{table}[t]
\centering
\caption{\textbf{Ablation Study.}
(Fig.~\ref{fig:method_overview}) We ablate the semantic proxy room (PR), geometry alignment module (GA), panorama generation (PG) and mesh completion (MC).
}
\label{table:high_level}
\resizebox{\linewidth}{!}{
\begin{tabu}{ccccccccc}
 \arrayrulecolor{black}
 \toprule
  \multirow{2}[2]{*}[0ex]{\shortstack{PR\\\colorsquare{m_red}}} & \multirow{2}[2]{*}[0ex]{\shortstack{GA\\~\colorsquare{m_green}}} & \multirow{2}[2]{*}[0ex]{\shortstack{PG\\\colorsquare{m_blue}}} & \multirow{2}[2]{*}[0ex]{\shortstack{MC\\\colorsquare{m_yellow}}} & \multicolumn{2}{c}{2D Metrics} & \multicolumn{3}{c}{User Study \small{(3D Mesh)}} \\
 \cmidrule(l{2pt}r{2pt}){5-6} \cmidrule(l{2pt}r{2pt}){7-9}
  &  &  & & CS $\uparrow$ & IS $\uparrow$ & 3DS $\uparrow$ & PA $\uparrow$ & PQ $\uparrow$ \\
 \cmidrule(r{4pt}){1-4} \cmidrule(l{4pt}){5-9}
 \xmark & \xmark & \xmark & \xmark & 27.0 & 5.76 & 1.78 & 1.14 & 2.33 \\
 \cmark & \xmark & \xmark & \xmark & 27.6 & 5.60 & 2.81 & 2.02 & 2.95 \\
 \cmark & \cmark & \xmark & \xmark & 27.8 & 5.64 & 3.34 & 4.34 & 3.45 \\
 \cmark & \cmark & \cmark & \xmark & \textbf{28.4} & \textbf{5.90} & 3.68 & 4.55 & 3.95 \\
\arrayrulecolor{black!10}\midrule\arrayrulecolor{black}
 \cmark & \cmark & \cmark & \cmark & \textbf{28.4} & 5.75 & \textbf{4.22} & \textbf{4.76} & \textbf{4.23} \\
 \bottomrule
\end{tabu}
}
\end{table}
The key technical components of our methods comprise the \emph{proxy room}~\colorsquare{m_red}, the \emph{geometry alignment module}~\colorsquare{m_green}, the \emph{panorama generation}~\colorsquare{m_blue}, and the final \emph{mesh completion}~\colorsquare{m_yellow}.
In Tab.~\ref{table:high_level}, we evaluate the influence of each component.
In the supplementary material, we provide further results about the adapter fine-tuning on HyperSim as well as visualizations of baseline methods of the ablation study.
\newpage
\parag{Adapter Fine-tuning on HyperSim Bounding Boxes.}
Fig.~\ref{fig:hypersim_comparison} presents a side-by-side qualitative evaluation of \name{}, illustrating its performance with and without fine-tuning on our 3D bounding box dataset derived from HyperSim.
A critical observation is that when \name{} uses adapters only trained on datasets of pixel-precise masks, it faces challenges in generating objects accurately within their designated bounding boxes.
Instead, it tends to create box-shaped objects that entirely fill the bounding boxes (Fig.~\ref{fig:hypersim_comparison}\textcolor{red}{a}).
However, when the adapters are fine-tuned using our dataset derived from HyperSim, \name{} demonstrates an improved ability to generate objects that appropriately fit within their corresponding bounding boxes (Fig.~\ref{fig:hypersim_comparison}\textcolor{red}{b}).

\parag{Consistent Style Panorama Generation.}
The participants in our user study expressed a strong preference for the panorama generation module~\colorsquare{m_blue}, noting its superiority across all three metrics compared to the incremental inpainting approach.
Incremental inpainting involves warping generated images to subsequent ones and autoregressively filling in unseen areas.
Since this inpainting is solely based on local context, it leads to gradual changes in style and challenges in achieving accurate loop closure~\cite{tang23neurips} (Fig.~\ref{fig:sem_control_mvdiffusion}\textcolor{red}{a}, \emph{white box}).
In contrast, our panorama generation module creates a complete 360$^\circ$ panorama, conditioned on both semantic and geometric 3D information in a unified manner.
This approach ensures a consistent style throughout the entire scene, %
effectively addressing the issues associated with incremental inpainting and resulting in a more cohesive and visually appealing overall scene (Fig.~\ref{fig:sem_control_mvdiffusion}\textcolor{red}{b}, \emph{bottom}).

\begin{figure}[t]
\includegraphics[width=\linewidth]{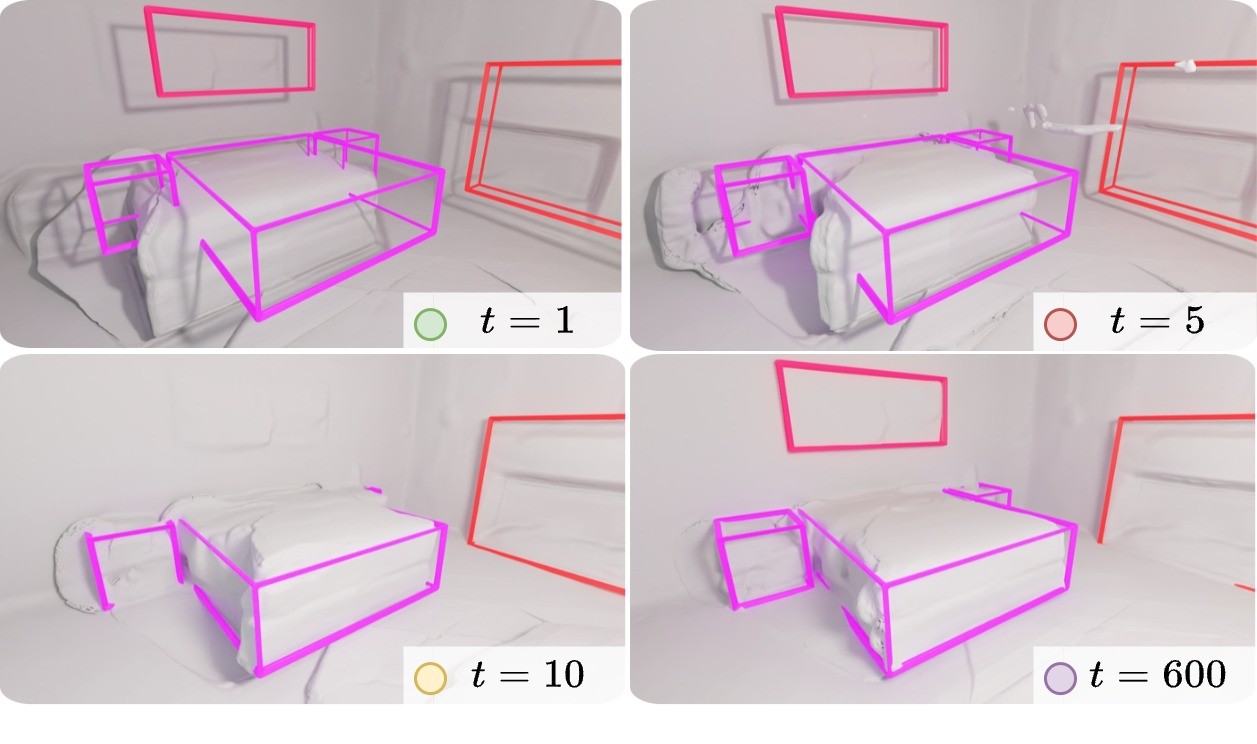}
\vspace{-28px}
\caption{\textbf{Geometry Alignment.}
Scale ambiguity leads to significant inaccuracies in MDEs (\colordot{m_green}).
In contrast, our proposed depth alignment module optimizes the alignment loss $\mathcal{L}$ to achieve strong alignment with the proxy room (\colordot{m_violet}).
\vspace{-11px}
}
\label{fig:depth_align_steps}
\end{figure}

\parag{Geometry Alignment with 3D Boxes.}
Estimating the exact scale of a 3D scene from a single RGB image is an ill-posed task, often leading to severe inaccuracies in even the most advanced metric depth estimators~\cite{bhat23arxiv}. As shown in Fig.~\ref{fig:depth_align_steps}, without iterative alignment, the estimated depth would lead to implausible geometry  ($t$$=$$1$).
By iteratively optimizing the 3D representation with additional geometric information of 3D bounding boxes of the proxy room layout, the final geometry is more plausible ($t$$=$$600$). 
The geometry alignment module is the core ingredient to achieve strong alignment with the proxy room, as evident by the substantial improvement of +2.32 PA in Tab.~\ref{table:high_level}.

\begin{figure}[t]
\centering
\includegraphics[width=\linewidth]{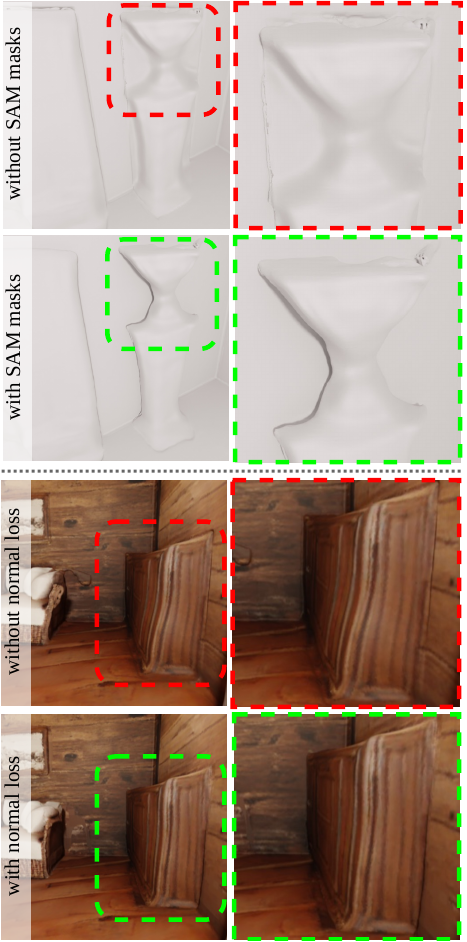}
\vspace{-10px}
\caption{\textbf{Ablation on Depth Alignment.}
Leveraging pixel-perfect SAM masks yields crisper 3D object boundaries (\emph{top}).
While the depth alignment loss aligns the frame with the 3D proxy room,
it can distort the object's surface as it fits the depth within its bounding box.
Addressing this issue, we introduce our normal preservation loss to maintain the object's original shape (\emph{bottom}).
\vspace{0px}
}
\label{fig:supp_depth}
\end{figure}

In Fig.~\ref{fig:supp_depth}, we present an additional qualitative ablation study focusing on the use of SAM masks and the normal preservation loss, both integral components of the depth alignment module.
We leverage SAM~\cite{kirillov23arxiv} to obtain pixel-precise instance masks for each generated object.
For pixels located within the rendered bounding box but outside the SAM mask, we assign the depth value $D_n$ to $D_f$.
As shown in Fig.~\ref{fig:supp_depth} (\emph{top}), including SAM masks leads to sharper 3D object boundaries, resulting in a more seamless integration into the 3D room mesh.
Although the depth alignment loss $\mathcal{L}_d$ effectively aligns the frame with the 3D proxy room, it may occasionally distort the surface of objects to fit them within their bounding boxes.
To counter this, we introduce the normal preservation loss $\mathcal{L}_n$, retaining the original shape of the objects (Fig.~\ref{fig:supp_depth}, \emph{bottom}).

\parag{High Quality Generation with Mesh Completion.}
Once the geometry alignment module has aligned the generated geometry with the layout, we can render control signals from the proxy room, including semantic, instance, and guidance depth maps, for selected new viewpoints that correspond with the created 3D mesh. 
This process enables us to fill in the missing parts of the mesh which remained unobserved after the panorama generation with localized inpainting,
and incorporate new content into the 3D structure, guided by these aligned control signals.
This yields notable enhancement in structural completeness (+0.54~3DS), culminating in the highest perceptual quality (4.23 PQ).

\vspace{-5px}
\section{Conclusion}
\vspace{-5pt}
In this paper, we proposed \name{}, a novel method to generate high-quality 3D room meshes given a textual description of the room style and a user-defined room layout outlined by 3D semantic bounding boxes.
When rendered to 2D, the geometric and semantic information is used to guide generative models to create texture and geometry aligned with the proxy room's layout. 
Naive guidance does not produce plausible meshes, thus, we introduced guided panorama generation and geometry alignment modules to ensure consistent style and plausible geometry. 
Finally, we introduced mesh post-processing and completion methods, improving the final generation quality substantially.
We believe our method marks a significant step forward in enabling the creation of large-scale 3D assets through user-defined control signals, greatly simplifying the process for end-users without any expert knowledge in 3D modeling software to design personalized scenes.

\parag{Acknowledgments.}
We would like to thank Xiaoliang Dai, Jia-Bin Huang, Jeff Liang and Matthew Yu for helpful feedback and discussions.

\clearpage

{
    \small
    \bibliographystyle{ieeenat_fullname}
    \bibliography{main}
}
\clearpage
\maketitlesupplementary

\textit{
In this supplementary material, we provide mode details of our technical components and experimental setup. %
Specifically, we provide further details about our user study and conclude with additional qualitative results.
For a comprehensive understanding, we suggest that reviewers watch our supplemental video, which includes detailed explanations and showcases videos of 3D room meshes in a variety of room types and styles.
}

\section{Adapter Fine-tuning on HyperSim}
\begin{figure}[t]
\centering
\includegraphics[width=\linewidth]{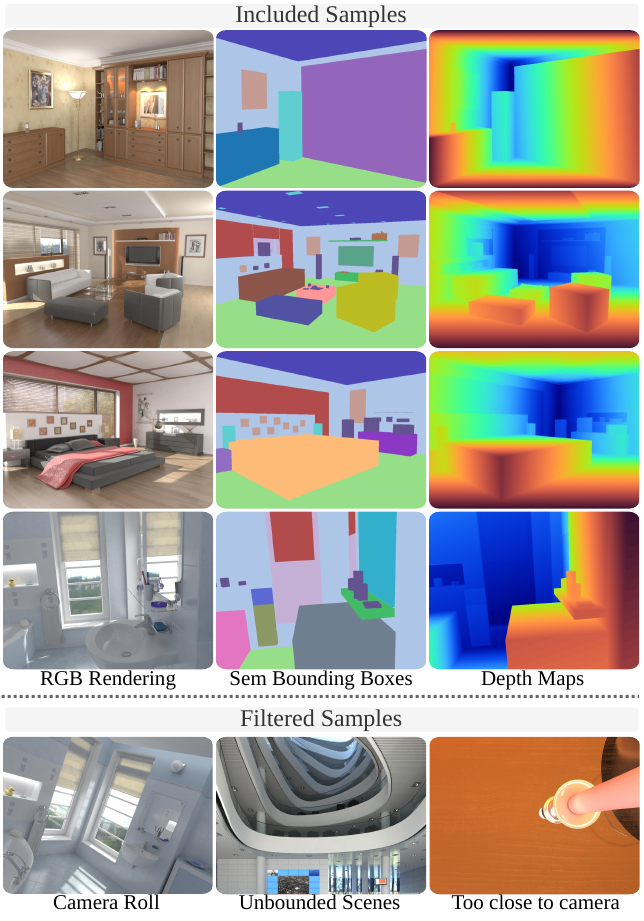}
\vspace{-20px}
\caption{\textbf{HyperSim Filtering.}
We show samples rendered from the HyperSim dataset which meet our quality requirements (\emph{top}).
Samples with camera roll, extreme far shots and close-ups are filtered out to better match \name{}'s use case (\emph{bottom}).
\vspace{-10px}
}
\label{fig:hypersim_filtering}
\end{figure}

\parag{Dataset Preparation and Implementation Details.}
The HyperSim dataset~\cite{roberts21iccv} offers 2D rendered images that include camera positions and aligned 3D semantic bounding boxes.
Using these camera poses, we project the semantic bounding boxes into 2D, creating depth maps and semantic maps of the bounding boxes (see Fig.~\ref{fig:hypersim_filtering}, \emph{top}).
However, we observed that some images from the HyperSim dataset do not meet our quality requirements (Fig.~\ref{fig:hypersim_filtering}, \emph{bottom}).
Therefore, we exclude images where the camera roll exceeds $\pm 8.6^\circ$, as our camera setup is typically parallel to the ground. Additionally, we discard images where a single semantic class covers more than half of the image area, as this often suggests the camera is too close to objects in the 3D scene or inside a bounding box.
Images with a maximum depth value over 15 meters are also removed, considering our focus on bounded indoor environments.
For the selected images, we calculate BLIPv2 captions.
We fine-tune the semantic and depth adapter independently on their respective maps.
We train for 5000 iterations on 8 A100 GPUs, with a batch size of 8 and a learning rate of 1e-7.

\parag{Qualitative Comparison.}
Fig.~\ref{fig:hypersim_comparison} presents a side-by-side qualitative evaluation of \name{}, illustrating its performance with and without fine-tuning on our 3D bounding box dataset derived from HyperSim.
A critical observation is that when \name{} uses adapters only trained on datasets of pixel-precise masks, it faces challenges in generating objects accurately within their designated bounding boxes.
Instead, it tends to create box-shaped objects that entirely fill the bounding boxes (Fig.~\ref{fig:hypersim_comparison}\textcolor{red}{a}).
However, when the adapters are fine-tuned using our dataset derived from HyperSim, \name{} demonstrates an improved ability to generate objects that appropriately fit within their bounding boxes (Fig.~\ref{fig:hypersim_comparison}\textcolor{red}{b}).

\section{Details on User Study}
\begin{figure*}[t]
\vspace{0px}
\centering
\includegraphics[width=\linewidth,trim={0 0cm 0 0cm},clip]{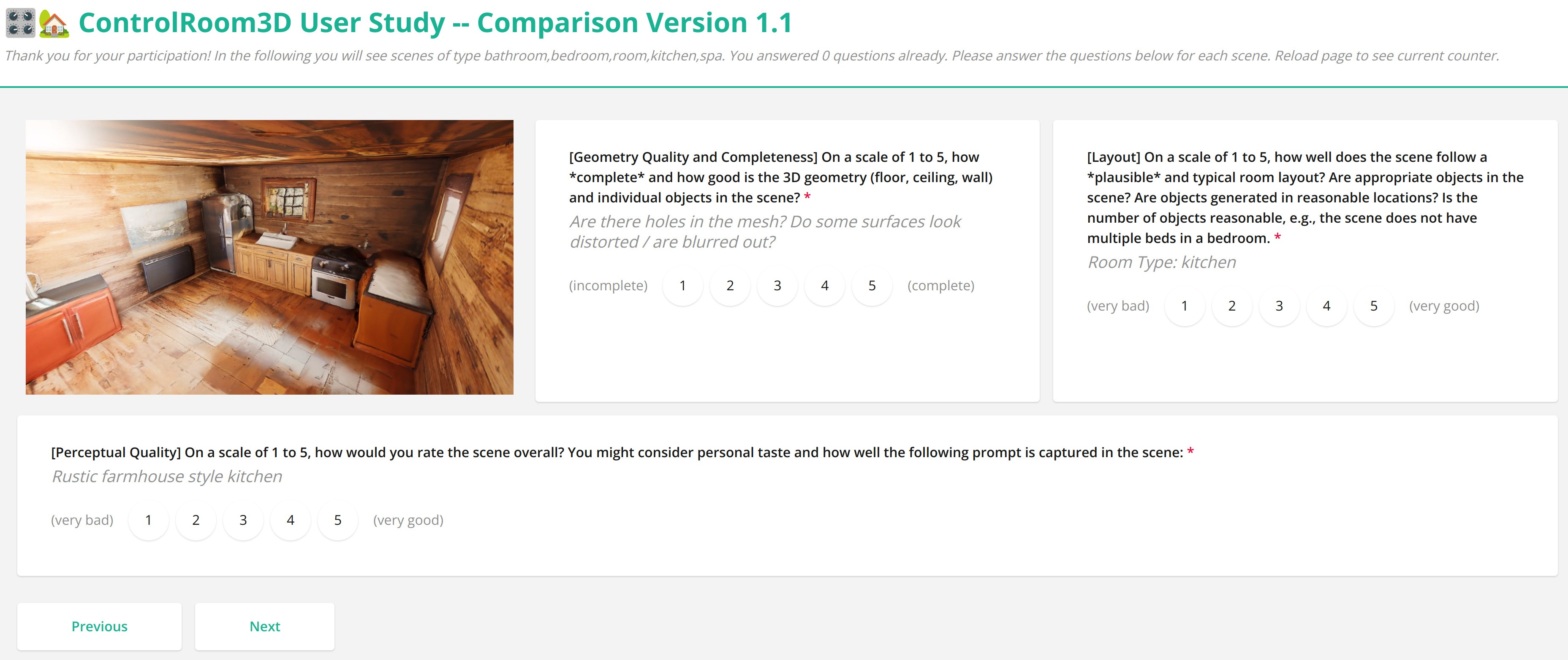}
\vspace{-10px}
\caption{\textbf{User Study Interface.}
We render 25sec long videos for each generated room showing all objects from multiple (challenging) angles.
We ask participants to rate these scenes on a scale of 1--5 with respect to the following dimensions: 3D structure completeness (3DS), proxy alignment to the 3D guidance (PA), generated room layout plausibility (LP), and overall perceptual quality (PQ).
\vspace{0px}
}
\label{fig:user_study}
\end{figure*}
\begin{figure*}[t]
\centering
    \centering
    \includegraphics[width=1.0\textwidth]{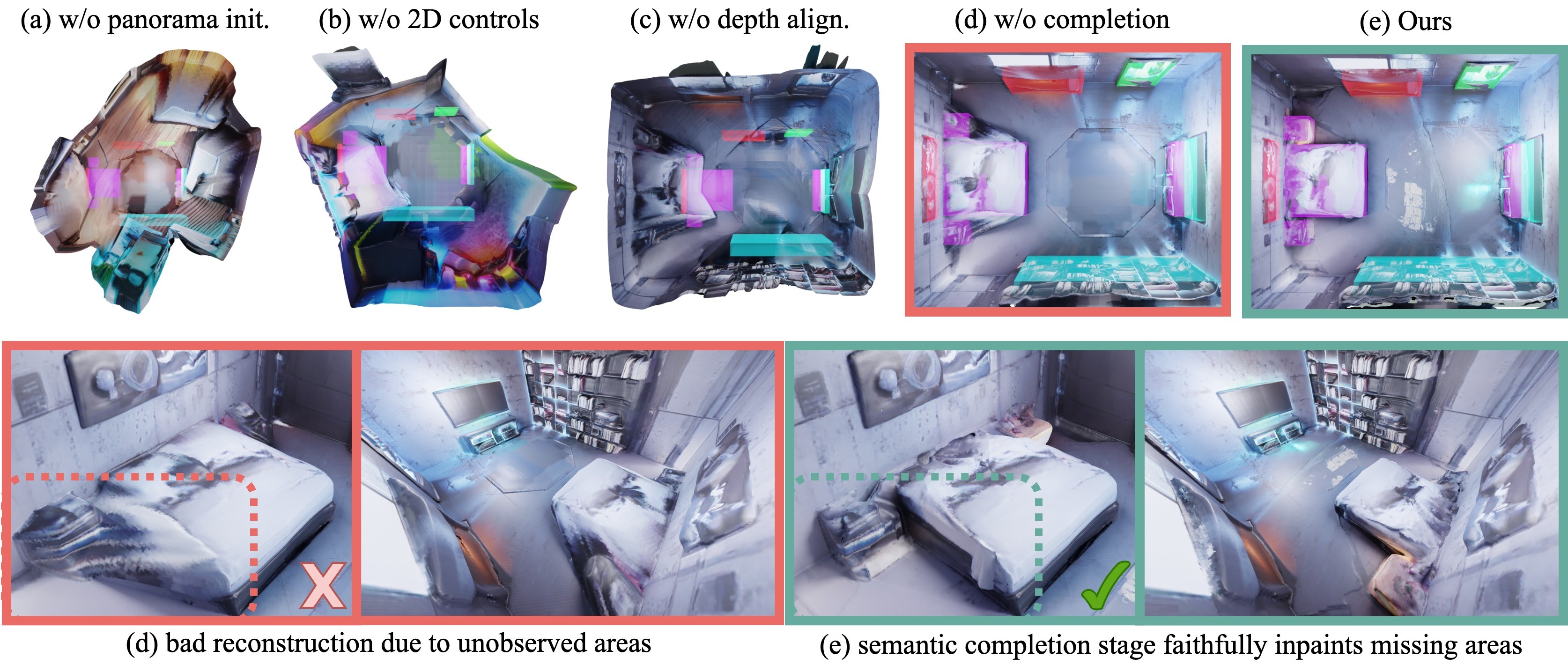}
    \vspace{-15px}
    \caption{\textbf{Qualitative Ablation on the key components of our method.}
    We notice that 2D guidance (c) plays a crucial role in creating a plausible room layout, \cf (a) -- (b).
    Nevertheless, the scene still lacks geometric alignment with the proxy room (\cf transparent bounding box overlays).
    Our depth alignment module accurately aligns the generated scene with the proxy room geometry (d).
    To address bad reconstruction artifacts due to unobserved areas (\emph{bottom left}), we employ our semantic completion stage to inpaint these missing regions.
    This results in complete 3D rooms without blurred-out sections (\emph{bottom right}).
    }
    \label{fig:qualitative_ablation}
\end{figure*}
We carried out two user studies in which we ask 12 participants to rate 3D scenes on a scale of 1--5 with respect to three qualitative dimensions.
Firstly, we compare our work with related work, \ie, Text2Room~\cite{hoellein23iccv} and an adaption of MVDiffusion~\cite{tang23neurips} for 3D room generation.
We ask all participants to rate each scene individually with respect to 3D structure (3DS), \ie, ``is the 3D mesh complete and not distorted?", layout plausibility (LP), \ie, ``does the room layout resemble a typical room layout of the specified type?" and overall perceptual quality (PQ).
We show the user study interface in Fig.~\ref{fig:user_study}.
Secondly, we conduct a user study for the ablation study in which we evaluate the influence of each technical component to our full model.
Here, we replace the question about layout plausibility with proxy alignment (PA), \ie, ``are objects generated within their corresponding bounding box?", and additionally superimpose 3D bounding boxes on the video to visualize the proxy room guidance. An example of this is included in the supplementary material.

\section{Further Qualitative Results}
\parag{Visualization of the Ablation Study.}
We provide an additional qualitative study regarding the key technical components of \name{} in Fig.~\ref{fig:qualitative_ablation} and show videos of an exemplary 3D room of each ablation study experiment in the supplementary material.
\newpage
\parag{Additional qualitative results of our main method.}
\begin{figure*}[th!]
\centering
    \centering
    \includegraphics[width=0.95\textwidth]{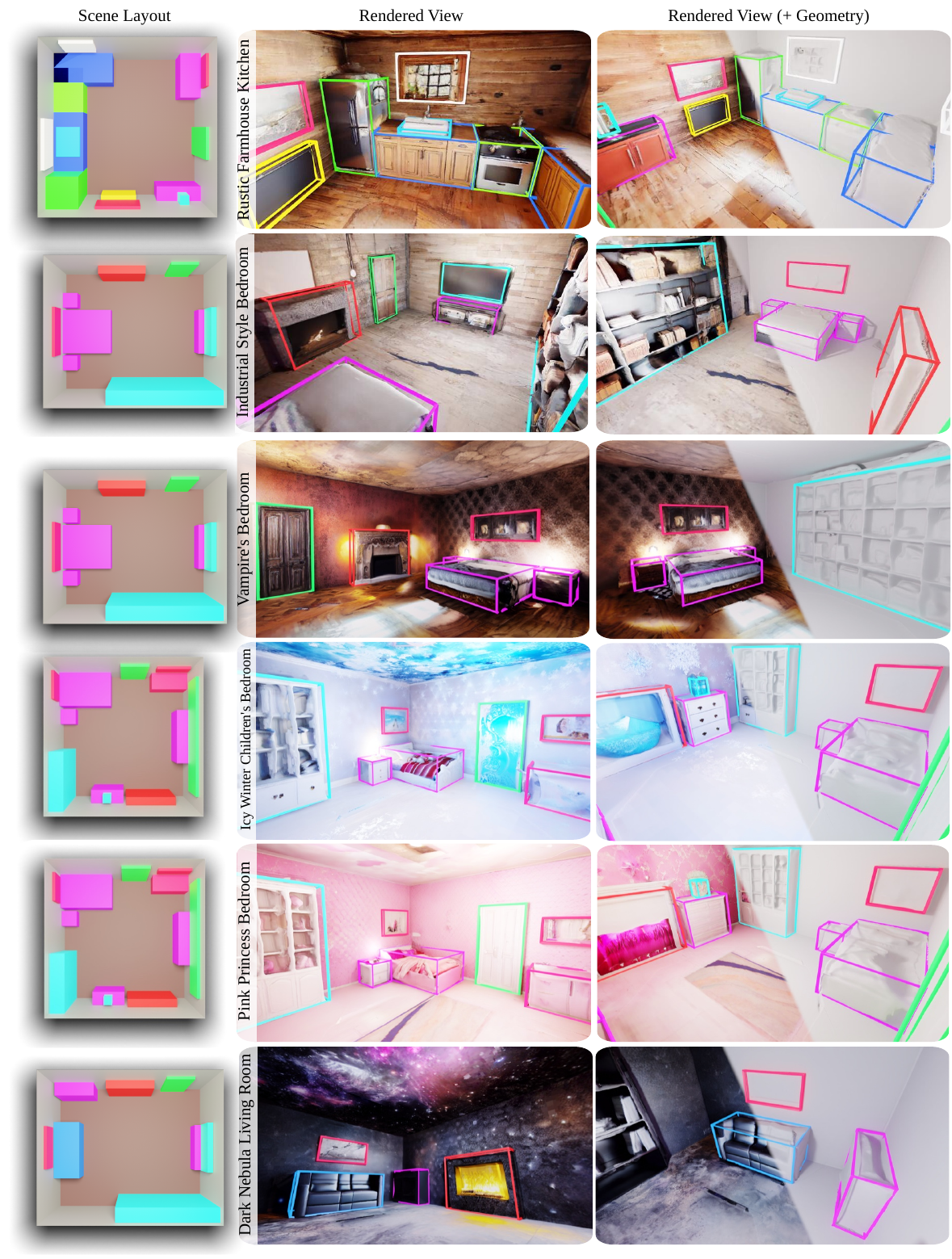}
    \vspace{-12px}
    \caption{\textbf{Further Qualitative Examples.}
    We show colored geometry renderings of our method.
    \name{} generates convincing geometries and textures given a user-defined room layout and textual style description.
    More qualitative videos can be found at the end of the explanatory video in the supplementary material.
    }
    \label{fig:qualitative_comparison2}
\end{figure*}
In Fig.~\ref{fig:qualitative_comparison2}, we show additional qualitative examples of \name{}.
Moreover, we recommend that reviewers watch our supplemental video explaining main technical component of \name{} as well as qualitative videos of 3D room meshes of various room types and styles.

\end{document}